\documentclass{article}
\usepackage{spconf,amsmath,graphicx,hyperref}
\usepackage{amsmath}   % For advanced math formatting
\usepackage{amssymb}   % For special math symbols
\usepackage{multirow}
\usepackage{bm}     
\usepackage{tabularx}
\usepackage{booktabs}
\usepackage{array}
\newcolumntype{Y}{>{\centering\arraybackslash}X} % 居中的自适应列% For bold math symbols (alternative to \mathbf)
\title{DrivingScene: A Multi-Task Online Feed-Forward 3D Gaussian Splatting Method for Dynamic Driving Scenes}
\name{
    Qirui Hou$^{1,2}$,
    Wenzhang Sun$^{3}$,
    Chang Zeng$^{3}$,
    Chunfeng Wang$^{3}$,
    Hao Li$^{3}$,
    Jianxun Cui$^{1,2}$\textsuperscript{*}\thanks{\textsuperscript{*}Corresponding Author} % 星号表示通讯作者
}

\address{
    $^{1}$ Harbin Institute of Technology, China  \\
    $^{2}$ Chongqing Research Institute of HIT, China \\
    $^{3}$ Li Auto, China
}
\begin{document}
\ninept
%
% \author{
%     % Author Name$^{\star \dagger}$ \qquad
%     Qirui Hou$^{\star}$ \qquad
%     Wenzhang Sun$^{\dagger}$ \qquad
%     Chang Zeng$^{\dagger}$ \qquad
%     Chunfeng Wang$^{\dagger}$ \qquad
%     Hao Li$^{\dagger}$ \qquad
%     Jianxun Cui$^{\star}$
% }

% \address{
%     $^{\star}$ Harbin Institute of Technology \\
%     $^{\dagger}$ Li Auto
% }
\maketitle
\begin{abstract}
% Real-time, high-fidelity reconstruction of dynamic driving scenes is challenged by 
% complex dynamics and sparse views, with prior methods struggling to balance quality and 
% efficiency. We proposed \mbox{DrivingScene}, an online, feedforward framework that 
% reconstructs 4D dynamic scenes from only two consecutive surround view images. Our multitask method represents scenes with 3D Gaussians and employs a  two-stage 
% training strategy. First, we train depth and Gaussian networks self-supervisedly to 
% establish a robust static scene prior. Second, with the static backbone frozen, we 
% train a residual flow network to predict temporal displacements for the Gaussians, 
% explicitly modeling dynamic elements. Experiments on nuScenes show that our image-only 
% method jointly produces high-quality depth, scene flow, and point clouds, significantly 
% outperforming state-of-the-art approaches in both dynamic reconstruction and  novel view 
% synthesis.
Real-time, high-fidelity reconstruction of dynamic driving scenes is challenged by complex dynamics and sparse views, with prior methods struggling to balance quality and efficiency. We propose DrivingScene, an online, feed-forward framework that reconstructs 4D dynamic scenes from only two consecutive surround-view images. Our key innovation is a lightweight residual flow network that predicts the non-rigid motion of dynamic objects per camera on top of a learned static scene prior, explicitly modeling dynamics via scene flow. We also introduce a coarse-to-fine training paradigm that circumvents the instabilities common to end-to-end approaches. Experiments on nuScenes dataset show our image-only method simultaneously generates high-quality depth, scene flow, and 3D Gaussian point clouds online, significantly outperforming state-of-the-art methods in both dynamic reconstruction and novel view synthesis.
\end{abstract}
\begin{keywords}
Autonomous Driving, Novel view Synthesis, Multi task Learning
\end{keywords}
\section{Introduction}
\label{sec:introduction}

Accurate, real-time 4D (3D space + time) environmental perception and reconstruction form the bedrock of safety and reliability for autonomous driving systems. Modern autonomous vehicles are typically equipped with multiple cameras for 360-degree surround-view perception. Compared to fusion-based approaches that rely on multi-modal sensors like LiDAR or RaDAR\cite{geiger2012we,caesar2020nuscenes,sun2020scalability}, vision-only methods\cite{philion2020lift,li2024bevformer} offer a more cost-effective and computationally efficient pathway for complex online perception tasks. However, reconstructing a large-scale, geometrically accurate, and photorealistic dynamic scene in real-time, solely from sparse and dynamic surround-view images, remains a significant and unresolved challenge.

The pursuit of higher reconstruction fidelity has seen tremendous success with neural rendering techniques like NeRF~\cite{mildenhall2021nerf} and 3DGS~\cite{kerbl20233d}. However, the majority of these methods, whether for static scenes like StreetGaussian~\cite{yan2024street}, DrivingGaussian~\cite{zhou2024drivinggaussian} or dynamic scenes like EmerNeRF~\cite{yang2023emernerf}, are bound by a per-scene optimization paradigm. This reliance on time-consuming offline training is incompatible with the real-time requirements of autonomous driving downstream tasks, necessitating a paradigm shift towards "feed-forward" reconstruction\cite{wang2021ibrnet,yu2021pixelnerf,wang2025vggt,xu20254dgt}. This online approach has matured for static scenes, with methods like pixelSplat~\cite{charatan2024pixelsplat} and MVSplat~\cite{chen2024mvsplat} demonstrating its viability, and culminating in works like DrivingForward~\cite{tian2025drivingforward} which successfully handle sparse driving contexts. Yet, their foundational static world assumption inevitably leads to severe artifacts when confronted with moving vehicles. To address this, methods like Driv3R~\cite{fei2024driv3r} have attempted to model dynamic scenes end-to-end. However, this monolithic design not only imposes a heavy computational burden but, more importantly, fails to explicitly decouple the inherently distinct static and dynamic components of a scene, leaving room for improvement in reconstruction detail and fidelity.

\begin{figure}
    \centering
    \includegraphics[width=\linewidth]{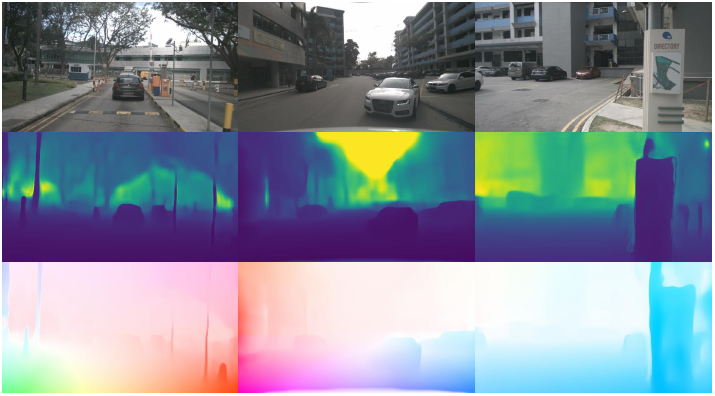}
    \caption{Example predictions by our method on nuScenes~\cite{caesar2020nuscenes}. Top to bottom: input image (one of the sequence), depth map and optical flow. Our model is fully self-supervised and can handle dynamic objects and occlusions explicitly.}
    \label{fig:example}
\vspace{-10pt}
\end{figure}
% \begin{figure*}
%     \centering
%     \includegraphics[width=\linewidth]{figures/pipeline.pdf}
%     \caption{Overview of DrivingScene. Given two sparse surround view frames from vehicle cameras, a depth network and a Gaussian network first cooperate to reconstruct a high-quality static scene. The key innovation of our method is a residual flow network, which precisely captures the non-rigid motion field between the frames. This motion is then applied to the geometric means of the static Gaussians, animating the scene into a dynamic 4D representation and enabling efficient, feed-forward reconstruction.}
%     \label{fig:visual}
% \end{figure*}
\begin{figure*}
    \centering
    \includegraphics[width=\linewidth]{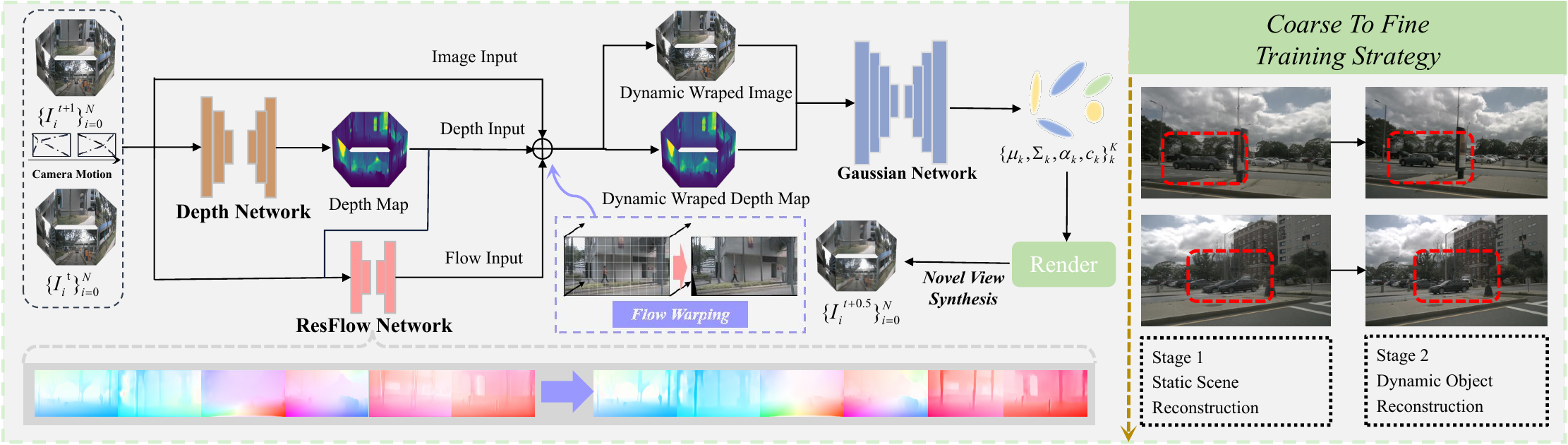}
    \caption{Overview of DrivingScene. Given two consecutive surround-view frames, our framework first predicts a static scene composed of 3D Gaussian primitives using a depth and a Gaussian parameter network. A residual flow network then computes the non-rigid motion field between the frames. This motion is combined with the rigid flow derived from ego-motion and applied as temporal displacements to the static Gaussians, resulting in a complete, dynamic 4D scene representation.}
    \label{fig:visual}
\vspace{-8pt}
\end{figure*}
To address these challenges, we introduce DrivingScene, an efficient online, feed-forward framework designed specifically for online dynamic driving scene reconstruction. The key to our approach is a two-stage, static-to-dynamic learning strategy that decouples the complex 4D reconstruction problem into two more tractable sub-tasks: robust static scene modeling and subsequent dynamic refinement. Specifically, in the first stage, we focus on training a network to learn a powerful static scene prior from large-scale data. This initial phase establishes a high-fidelity and geometrically consistent foundation for the static components of the world, such as buildings and road infrastructure. Upon convergence, we freeze this static backbone and introduce a lightweight residual flow network\cite{ilg2017flownet,yin2018geonet}. This network is uniquely trained to predict only the non-rigid motion residuals corresponding to independently moving objects, rather than the entire motion field. This progressive, static-to-dynamic paradigm offers several advantages: it effectively circumvents the training instabilities common to monolithic end-to-end approaches, and by decomposing the motion, it allows our model to generate temporally coherent and detailed high-fidelity dynamic scenes with computational efficiency necessary for real-time performance.

The main contributions of this paper are summarized as follows: 1) We proposed DrivingScene, an online, feed-forward framework that achieves state-of-the-art 4D dynamic scene reconstruction from only two surround view images and generates valuable intermediate representations, it operates in real-time and trained entirely with self-supervised objectives. 2) We design a residual flow network with a hybrid-shared architecture. It features a shared backbone to learn a generalized motion prior and lightweight, per-camera heads to adapt to varying camera extrinsics and intrinsics , which keeps consistent scale prediction and computational efficiency across all views. 3) We introduce a coarse-to-fine, two-stage training paradigm. In Stage 1, DrivingScene learns a robust static scene prior. In Stage 2, with the static backbone frozen, a residual flow network is trained to refine the scene by modeling only the non-rigid motion of dynamic objects, ensuring both training stability and high-fidelity results.

\section{Methodology}
\label{sec:method}
% \subsection{Overview}
% \label{ssec:overview}
% We introduce DrivingScene, an online, feedforward framework for reconstructing spatio-temporally consistent 4D dynamic scenes from two consecutive, sparse surround view images. Figure~\ref{fig:visual} illustrates the overall framework of \textbf{DrivingScene}. At its core is a synergistic multitask architecture that represents the scene with 3D Gaussian primitives and comprises three key modules: a depth estimation network, a Gaussian parameter network, and a residual flow network.
We introduce DrivingScene, an online, feedforward framework for reconstructing spatio-temporally consistent 4D dynamic scenes from two consecutive, sparse surround view images. Figure~\ref{fig:visual} illustrates the overall framework of \textbf{DrivingScene}. 
% To effectively learn both static and dynamic scene properties, we devise a principled two-stage training strategy. In the first stage, we establish a robust geometric and appearance prior for the static scene. In the second, we introduce a residual flow network to explicitly model the motion of dynamic elements. This progressive approach enables DrivingScene to learn a powerful scene prior and perform real-time, high-fidelity reconstruction at inference.
% To effectively learn both static and dynamic scene properties, we devise a principled, coarse-to-fine training paradigm. The first stage provides us with a robust geometric and appearance prior, effectively capturing the rigid layout of the scene. However, this representation is inherently static and ignores the prevalent motion of dynamic objects. While a generic optical flow network could be employed to model the entire motion field, such an approach would fail to leverage the well-constrained nature of the rigid scene components that our first stage has already modeled.
To effectively learn both static and dynamic scene properties, we devise a coarse-to-fine training paradigm. The first stage provides a robust prior for the scene's rigid layout but ignores dynamic motion. Instead of a generic flow network, which would disregard these learned rigid constraints, we introduce the residual flow network, which is trained specifically to predict only the residual, non-rigid motion of dynamic objects on top of the frozen static backbone. This progressive, static-to-dynamic approach enables DrivingScene to explicitly model dynamics via scene flow and perform online, high-fidelity reconstruction.
\subsection{Static scene geometry and appearance modeling}
\label{ssec:static_scene_rec}
We ground our scene representation in 3D Gaussian Splatting (3DGS), which explicitly models a scene with a set of Gaussian primitives $\mathcal{G} = \{ G_k = \{ \boldsymbol{\mu}_k, \boldsymbol{\Sigma}_k, \alpha_k, \mathbf{c}_k \} \}_{k=1}^K$, parameterized by a 3D mean $\boldsymbol{\mu}_k$, a covariance matrix $\boldsymbol{\Sigma}_k$, an opacity  $\alpha_k$, and Spherical Harmonic (SH) coefficients $\mathbf{c}_k$. To enable feed forward inference, we design a depth network $D$, and a Gaussian parameter network $P$, to directly predict these properties from images.

Given a pair of consecutive surround view image sets and their poses, the depth network $D$ first predicts a per-pixel depth map for each image, which provides the 3D means ($\boldsymbol{\mu}_k$) for the Gaussian primitives. Subsequently, the network $P$ takes image and depth features as input to infer the remaining attributes.

The Gaussian primitives predicted from each of the six camera views are transformed into a common world coordinate system using the known extrinsic parameters. These individual point clouds are then concatenated to form a single unified scene representation. In this feed-forward paradigm, we do not perform explicit de-duplication or fusion in 3D space. Instead, we rely on the differentiable renderer to handle potential redundancies and inconsistencies during the view synthesis process, where Gaussians that are occluded or inconsistent with the target view will naturally contribute minimally to the final rendered pixel color. 

\subsection{Dynamic modeling via residual scene flow}
\label{ssec:dynamic_modeling}
\begin{figure}
    \centering
    \includegraphics[width=\linewidth]{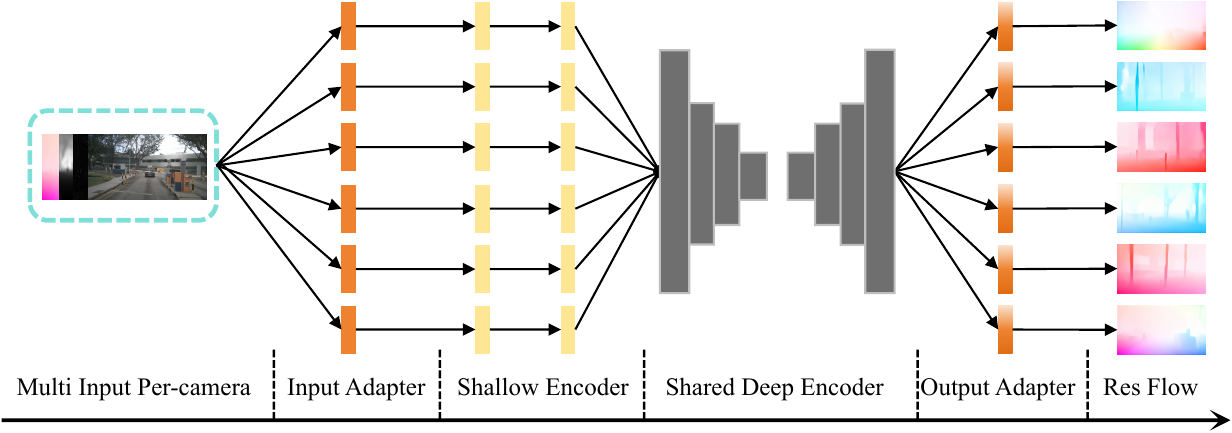}
    \caption{The architecture of residual flow network}
    \label{fig:network_arch}
\vspace{-15pt}
\end{figure}
\begin{figure*}{t}
    \centering
    \includegraphics[width=\linewidth]{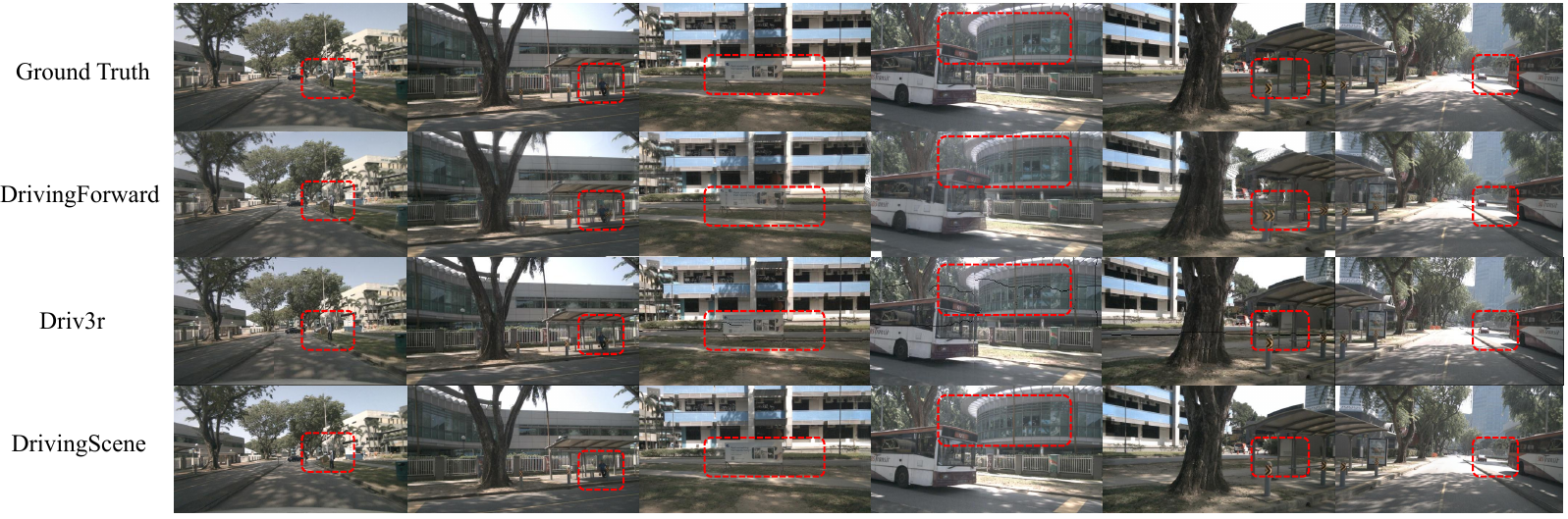}
    \caption{Qualitative results of  surrounding views. Details from surrounding views are present for easy comparison.}
    \label{fig:result}
\vspace{-15pt}
\end{figure*}
The static model established in Stage 1 is inherently incapable of capturing independently moving objects. To model these dynamics, we introduce a residual flow network, $R$. The central principle is to decompose the total motion field into a rigid component $\mathbf{F}_{\text{rigid}}$ and a non-rigid residual component $\mathbf{F}_{\text{residual}}$. This allows the network to focus on learning challenging, object-specific motion.

To achieve this efficiently online across multiple views, we introduce a  hybrid architecture for $R$ (as depicted in Figure~\ref{fig:network_arch}). It follows a coarse-to-fine principle, featuring a shared deep encoder backbone to extract a multiscale pyramid of generic motion features and dedicated per-camera pyramidal decoders. During decoding, the flow is iteratively refined from the lowest resolution upward, with each level's prediction serving as an initial estimate for the next. This pyramidal refinement strategy is critical for handling large displacements, while the hybrid design ensures consistent scale and a compact parameter footprint.

The process is as follows: we first compute the rigid flow field $\mathbf{F}_{\text{rigid}}$ using the predicted depth and known camera poses. The network $R$ then takes a rich set of inputs, including the warped source image, the target image, and the rigid flow, to predict the final residual component $\mathbf{F}_{\text{residual}}$. The complete motion field, $\mathbf{F}_{\text{total}} = \mathbf{F}_{\text{rigid}} + \mathbf{F}_{\text{residual}}$, is then applied to the means of Gaussian primitives to model their temporal evolution.

\subsection{Two-stage training and objectives}
\label{ssec:two_stage_training}
We propose a two-stage coarse-to-fine training strategy that decouples the learning of static and dynamic scene properties, mitigating the challenges of joint end-to-end optimization.

In the first stage, we exclusively train the depth $D$ and Gaussian parameter $P$ networks. The training is guided by a self-supervised composite loss function $\mathcal{L}_{\text{stage1}}$, which combines geometric and rendering objectives:
$$
\mathcal{L}_{\text{stage1}} = \lambda_{\text{loc}}\mathcal{L}_{\text{loc}} + \lambda_{\text{smooth}}\mathcal{L}_{\text{smooth}} + \lambda_{\text{render}}\mathcal{L}_{\text{render}}
$$
Here, the geometric loss, $\mathcal{L}_{\text{loc}}$, introduced from DrivingForward\cite{tian2025drivingforward}, enforces multiview consistency through a photometric reprojection objective. The smoothness loss $\mathcal{L}_{\text{smooth}}$, is a regularization term that penalizes large gradients in the disparity map. Finally, the rendering loss  $\mathcal{L}_{\text{render}}$, ensures visual fidelity by minimizing the difference between the rendered image $I_{\text{render}}$ and the ground truth image $I_{\text{gt}}$ using a combination of L2  photometric loss and perceptual LPIPS losses~\cite{zhang2018unreasonable},  with the weight $\lambda_{\text{p}}$ set to 0.05:
$$
\mathcal{L}_{\text{render}} = \mathcal{L}_{\text{L2}}(I_{\text{render}}, I_{\text{gt}}) + \lambda_{\text{p}}\mathcal{L}_{\text{LPIPS}}(I_{\text{render}}, I_{\text{gt}})
$$

Upon convergence of the static model, we freeze the weights of $D$ and $P$ and exclusively train the residual flow network $R$. The total loss $\mathcal{L}_{\text{stage2}}$ is a weighted sum of three self-supervised components:
$$
\mathcal{L}_{\text{stage2}} = \lambda_{\text{warp}}\mathcal{L}_{\text{warp}} + \lambda_{\text{consist}}\mathcal{L}_{\text{consist}} + \lambda_{\text{render}}\mathcal{L}_{\text{render}}
$$
The flow consistency loss $\mathcal{L}_{\text{consist}}$ provides geometric regularization through a forward-backward check. The Gaussian rendering loss $\mathcal{L}_{\text{render}}$ uses the same formulation as in Stage 1 to provide end-to-end supervision. The flow warping loss $\mathcal{L}_{\text{warp}}$ enforces photometric consistency on the warped image $\hat{I}_{t+1} = W(I_t, \mathbf{F}_{\text{total}})$. It is a composite objective combining three distinct error metrics:
$$
\mathcal{L}_{\text{warp}} = \mathcal{L}_{\text{L1}}(I_{\text{t+1}}, \hat{I}_{t+1}) + \lambda_{\text{s}}\mathcal{L}_{\text{SSIM}}(I_{\text{t+1}}, \hat{I}_{t+1}) + \lambda_{\text{wp}}\mathcal{L}_{\text{LPIPS}}(I_{\text{t+1}}, \hat{I}_{t+1})
$$
where $\mathcal{L}_{\text{L1}}$, $\mathcal{L}_{\text{SSIM}}$, and $\mathcal{L}_{\text{LPIPS}}$ denote the L1 photometric loss, the Structural Similarity (SSIM) loss~\cite{wang2004image}, and the perceptual LPIPS loss, with weights set to $\lambda_{\text{s}}=0.1$ and $\lambda_{\text{wp}}=0.05$, respectively.

\section{Experiments}

\subsection{Experimental setup}
\label{ssec:setup}
Our model is implemented in PyTorch and trained on NVIDIA RTX5090 GPUs (32GB). We use the Adam optimizer with a learning rate of $1 \times 10^{-4}$ and a batch size of 1. Our two-stage training proceeds as follows: Stage 1 (6 epochs) uses loss weights $\lambda_{\text{render}}=0.01$, $\lambda_{\text{loc}}=0.1$, and $\lambda_{\text{smooth}}=0.001$. Stage 2 (6 epochs)  uses weights $\lambda_{\text{render}}=0.01$, $\lambda_{\text{consist}}=10^{-5}$, and $\lambda_{\text{warp}}=0.02$. We evaluate on the official split of the nuScenes dataset (700/150 scenes) at $352 \times 640$ resolution. The primary task is novel view synthesis of the intermediate temporal frame between two keyframes, evaluated using PSNR, SSIM, and LPIPS.

% \subsection{Baselines}
% \label{ssec:baselines}
We compare against leading online, feed-forward methods. Our primary baseline is DrivingForward~\cite{tian2025drivingforward}, a static reconstruction method whose limitations with dynamic objects we directly address. We also provide extensive comparisons against the dynamic method Driv3R~\cite{fei2024driv3r}, other static approaches (DepthSplat~\cite{xu2025depthsplat}, MVSplat), and the per-scene optimization method StreetGaussian, aligning our setup with their protocols for a fair comparison.

\subsection{Quantitative and qualitative comparison}
\label{ssec:main_results}
\label{sec:experiments}
% \begin{figure*}{t}
%     \centering
%     \includegraphics[width=\linewidth]{figures/result_image.pdf}
%     \caption{Qualitative results of  surrounding views. Details from surrounding views are present for easy comparison.}
%     \vspace{-6pt}
%     \label{fig:result}

% \end{figure*}
\begin{table}
\centering
\caption{Quantitative comparison for novel view synthesis on the nuScenes validation set.}
\label{tab:main_results}
\begin{tabularx}{\linewidth}{@{}l Y Y Y@{}} % 使用 \linewidth 适应单栏宽度
\toprule
Method & PSNR $\uparrow$ & SSIM $\uparrow$ & LPIPS $\downarrow$ \\
\midrule
% Feed-forward Static Methods
MVSplat &22.83& 0.629 & 0.327 \\
DepthSplat& 24.21 & 0.732 & 0.271 \\
StreetGaussian& 25.59 & 0.765 & 0.212 \\
DrivingForward & 26.06 & 0.781 & 0.215 \\
% \midrule
% Feed-forward Dynamic Method
Driv3R& 26.10 & 0.808 & \textbf{0.084} \\
% \midrule
% % Per-scene Optimization Method
% StreetGaussian& 25.59 & 0.765 & 0.212 \\
\midrule
\textbf{DrivingScene} & \textbf{28.76} & \textbf{0.895} & 0.113 \\
\bottomrule
\vspace{-15pt}
\end{tabularx}
\end{table}
Quantitative results for  novel view synthesis are presented in Table~\ref{tab:main_results}. \mbox{DrivingScene} achieves state-of-the-art performance, outperforming all feed-forward baselines across all metrics. This demonstrates the superior quality of our 4D reconstructions.

The qualitative comparisons in Figure~\ref{fig:result} further highlight the advantages of our approach. In particular, the comparison with \mbox{DrivingForward} showcases the critical importance of our dynamic modeling. While \mbox{DrivingForward} achieves strong results in static parts of the scene, its static assumption leads to significant ghosting and blurring artifacts for moving objects, such as vehicles and pedestrians. \mbox{DrivingScene} effectively resolves these dynamic elements, producing sharp and temporally consistent reconstructions that faithfully capture the scene's motion. Compared to \mbox{Driv3R}, our method demonstrates superior fine-grained detail reconstruction and overall visual fidelity.

% \subsection{Analysis of Multi-Task Outputs}
% \label{ssec:multitask_analysis}
\begin{table}[t]
\centering
\caption{Quantitative comparison for depth comprasion}
\label{tab:depth_results}
\begin{tabularx}{\linewidth}{@{}l Y Y Y@{}} % 使用 \linewidth 适应单栏宽度
\toprule
Method & Abs Rel  $\downarrow$& Sq Rel  $\downarrow$& RMSE $\downarrow$\\
\midrule
% \midrule
% Feed-forward Dynamic Method
Driv3R& 0.234 & 2.279 & 7.298\\
% \midrule
\textbf{DrivingScene} & \textbf{0.227} & \textbf{2.195} & \textbf{7.254}\\
\bottomrule
\vspace{-15pt}
\end{tabularx}
\end{table}
% \begin{figure}
%     \centering
%     \includegraphics[width=\linewidth]{figures/depth_compare.png}
%     \caption{The comparison of depth maps with Driv3r}
%     \label{fig:depth_comparison}
% \end{figure}
A key advantage of our framework is the generation of high-quality intermediate representations. We compare our predicted depth maps with \mbox{Driv3R}\cite{fei2024driv3r} in Table~\ref{tab:depth_results}. The results show that our method produces more accurate and geometrically coherent depth, validating the effectiveness of our explicit, multitask prediction approach. This superior geometric understanding is a key factor contributing to our higher rendering fidelity.
\begin{figure}
    \centering
    \includegraphics[width=\linewidth]{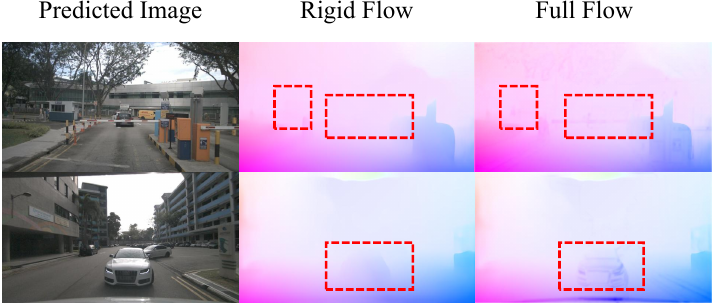}
    \caption{The comparison of rigid flow with full flow}
    \label{fig:flow_vis}
\vspace{-10pt}
\end{figure}
Furthermore, we visualize the decomposed flow fields in Figure~\ref{fig:flow_vis}. The rigid flow component correctly captures the global scene motion induced by the ego-vehicle, while the learned residual flow successfully isolates and highlights non-rigidly moving objects. This provides clear evidence for the efficacy of our residual motion modeling strategy.

\subsection{Efficiency analysis}
\label{ssec:efficiency_analysis}
\begin{table}[t!] % Using [t!] encourages LaTeX to place it at the top
\centering
\caption{Efficiency analysis}
\label{tab:efficiency}
\small
% The tabular* environment forces the table to a specific width (\linewidth).
% The @{\extracolsep{\fill}} command automatically adds the necessary space 
% between columns to fill that width.
\begin{tabular*}{\linewidth}{@{\extracolsep{\fill}}lcccc@{}}
\toprule
% & \multicolumn{2}{c}{Inference} & \multicolumn{2}{c}{Training} \\
\multirow{3}{*}{Method} & \multicolumn{2}{c}{Inference} & \multicolumn{2}{c}{Training} \\
\cmidrule(lr){2-3} \cmidrule(lr){4-5}
& Time $\downarrow$ & VRAM $\downarrow$ & Time $\downarrow$ & VRAM $\downarrow$ \\
% & (Hours) & (GB) & (GB)& (GB) \\
\midrule
DrivingForward & 0.34S & 7.58GB & \textbf{$\approx$ 3 days} & 40.0GB \\
Driv3R & 0.71s & \textbf{5.04GB} & $\approx$ 7.5days & 175.5GB \\
\midrule
\textbf{DrivingScene} & \textbf{0.21S} & 6.48GB & $\approx$ 5 days & \textbf{27.3GB} \\
\bottomrule
\vspace{-15pt}
\end{tabular*}
\end{table}

\begin{table}[h]
\centering
\caption{Model complexity comparison.}
\label{tab:parameters}
\small
\begin{tabular*}{\linewidth}{@{\extracolsep{\fill}}lccc@{}}
\toprule
Method& DrivingForward & Driv3R & \textbf{DrivingScene} \\
\midrule
Params $\downarrow$ & 0.173GB & 2.512GB & \textbf{0.117GB} \\
\bottomrule
\vspace{-20pt}
\end{tabular*}
\end{table}

We evaluated the computational efficiency of our method against \mbox{DrivingForward} and \mbox{Driv3R} in terms of training time, inference speed, and GPU memory consumption. As detailed in Table~\ref{tab:efficiency}, when synthesizing a full surround-view scene (6 images at $352 \times 640$), our method not only achieves a faster inference frame rate. The reported memory usage further confirms that our approach is more resource-efficient during both training and inference. Furthermore, as shown in Table~\ref{tab:parameters}, DrivingScene maintains a compact model size with significantly fewer parameters compared to \mbox{Driv3R} and DrivingForward. This highlights the efficiency of our hybrid-shared architecture and residual learning approach.
\subsection{Ablation studies}
\label{ssec:ablation_studies}
To systematically validate the key design choices of our method, we conduct a series of ablation studies. The results are summarized in Table~\ref{tab:ablation}.

\begin{table}[t]
\centering
\caption{Ablation studies on the key components of our method. }
\label{tab:ablation}
\begin{tabularx}{\linewidth}{@{}l Y Y Y@{}} % 使用 \linewidth 适应单栏宽度
\toprule
Configuration & PSNR $\uparrow$ & SSIM $\uparrow$ & LPIPS $\downarrow$ \\
\midrule
\textbf{Full Model} & \textbf{28.76} & \textbf{0.895} & \textbf{0.113} \\
\midrule
1. w/o Residual Flow & 26.40 & 0.780 & 0.201 \\
2. Single-Stage Training & 13.69 & 0.334 & 0.731 \\
3. w/o Flow Warping Loss & 27.32 & 0.872 & 0.145 \\
\bottomrule
\vspace{-20pt}
\end{tabularx}
\end{table}

\textbf{Efficacy of Residual Flow.} To verify the necessity of our dynamic modeling, we train a static-only variant of our model by disabling the residual flow network. This configuration is conceptually similar to the DrivingForward framework. The significant performance drop observed in the results confirms that explicitly modeling scene dynamics via our residual flow strategy is crucial for high-quality reconstruction in realistic driving scenarios.

\textbf{Efficacy of Two-Stage Training.} We compare our two-stage paradigm with a single-stage, end-to-end training alternative, where all loss functions are activated from the beginning. This joint training approach leads to a substantial degradation in performance. We observe that it impairs the model's ability to learn scale-aware geometry, underscoring the importance of establishing a robust static prior before refining with dynamic information.

\textbf{Efficacy of Flow Warping Loss.} Finally, we investigate the contribution of flow warping loss $\mathcal{L}_{\text{warp}}$ by removing it from the Stage 2 objective. The results show a noticeable decline in rendering quality, confirming that this loss provides a critical supervisory signal that tightly couples our motion estimation with the final rendering task, thereby enhancing multitask consistency.
\section{Conclusion}
\label{sec:conclusion}
% In this paper, we introduced \mbox{DrivingScene}, an online, feed-forward framework for high-fidelity 4D reconstruction of dynamic driving scenes. Our key innovation is a  two-stage, static-to-dynamic training strategy that decouples the learning of static geometry from dynamic motion, proving to be both effective and stable. \mbox{DrivingScene} surpasses key baselines like \mbox{DrivingForward} and \mbox{Driv3R} in rendering quality and computational efficiency, while concurrently generating high-quality intermediate outputs like depth and scene flow. 

% While {DrivingScene} demonstrates significant progress, we acknowledge several limitations that suggest avenues for future research. Our current dynamic model, which applies a single translational vector to each Gaussian mean, may be insufficient for modeling complex non-rigid deformations (e.g., pedestrians) or topological changes (e.g., an opening car door). Future work could explore integrating information over longer temporal windows to enhance robustness, or adopting more expressive, per-Gaussian deformation models to handle a wider range of dynamic phenomena. Nevertheless, our work represents a significant step towards enabling online, high-fidelity 4D environmental perception for autonomous systems.
In this paper, we introduced \mbox{DrivingScene}, an online, feed-forward framework for high-fidelity 4D reconstruction of dynamic driving scenes. Our key innovation is a  two-stage, static-to-dynamic training strategy that decouples the learning of static geometry from dynamic motion, proving to be both effective and stable. \mbox{DrivingScene} surpasses key baselines like \mbox{DrivingForward} and \mbox{Driv3R} in rendering quality and computational efficiency, while concurrently generating high-quality intermediate outputs like depth and scene flow. While \mbox{DrivingScene} shows significant progress, future work could explore integrating information over longer temporal windows to enhance robustness, or adopting more expressive, per-Gaussian deformation models to handle a wider range of dynamic phenomena.
% Nevertheless, our work represents a significant step towards enabling real-time, high-fidelity 4D environmental perception for autonomous systems.
% \section{Acknowlegdgment}
% \label{sec:conclusion}
% This research is funded by the Chongqing Natural Science Foundation Innovation and Development Joint Fund (Changan Automobile) (Grant No. CSTB2024NSCQ-LZX0157)
\bibliographystyle{IEEEbib}
\bibliography{strings,refs}

\end{document}